\documentclass[10pt,twocolumn,letterpaper]{article}

\usepackage{wacv}
\usepackage{times}
\usepackage{epsfig}
\usepackage{graphicx}
\usepackage{amsmath}
\usepackage{amssymb}
\usepackage{subfig}
\usepackage{graphicx}
\usepackage{algorithm}
\usepackage{algpseudocode}

\wacvfinalcopy 

\def\wacvPaperID{28} 

\ifwacvfinal\fi
\begin{document}

\title{Human and Sheep Facial Landmarks Localisation \\by Triplet Interpolated Features}

\author{Heng Yang*, Renqiao Zhang* and Peter Robinson \\
University of Cambridge\\
{\tt\small {hy306, rz264, pr10}@cam.ac.uk}
\thanks{* indicates authors contribute equally.}
}
\renewcommand\footnotemark{}
\renewcommand\footnoterule{}
\maketitle
\ifwacvfinal\thispagestyle{empty}\fi

\begin{abstract}
In this paper we present a method for localisation of facial landmarks on human and sheep. We introduce a new feature extraction scheme called  triplet-interpolated feature used at each iteration of the cascaded shape regression framework. It is able to extract features from similar semantic location given an estimated shape, even when head pose variations are large and the facial landmarks are very sparsely distributed. Furthermore, we study the impact of training data imbalance on model performance and propose a training sample augmentation scheme that produces more initialisations for training samples from the minority. More specifically, the augmentation number for a training sample is made to be negatively correlated to the value of the fitted probability density function at the sample's position.
We evaluate the proposed scheme on both human and sheep facial landmarks localisation. On the benchmark 300w human face dataset, we demonstrate the benefits of our proposed methods and show very competitive performance when comparing to other methods. On a newly created sheep face dataset, we get very good performance despite the fact that we only have a limited number of training samples and a set of sparse landmarks are annotated. 
\end{abstract}

\section{Introduction}
Many computer vision applications require localisation of a set of landmarks for the purpose of fine-grained recognition. For example, joint localisation in human pose estimation \cite{shotton2013real}, part localisation for bird \cite{berg2014birdsnap} and dog \cite{liu2012dog} breed recognition. It is of interest to localise facial landmarks for animals and humans, given the fact that their faces hold rich information such as identity, expression, health conditions, etc.  
\begin{figure}[!t]
 \centering
\includegraphics[width=0.5\textwidth,height=0.2\textwidth]{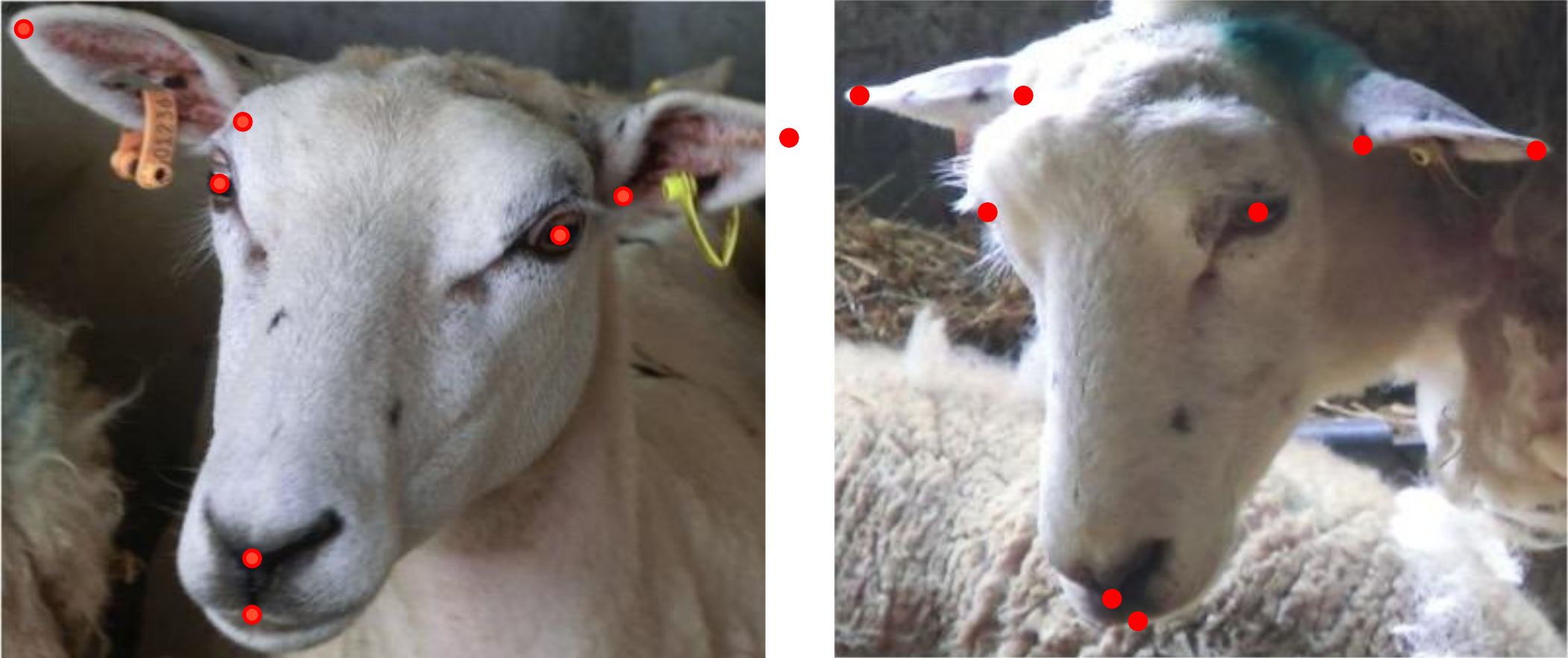}
\caption{Normal sheep (left) vs. sheep in pain (right). The red landmarks are associated with distinguishable patterns that we intend to localise. }
\label{fig:sheepinpain}
\end{figure}
In this paper, we are interested in localising sheep and human facial landmarks for real applications. Sheep facial landmark localisation is new in computer vision field and has very promising potential in animal welfare. Compared to other animals,  sheep have less intricate facial muscles and thus do not appear to have a wide array of facial expressions. However, researchers have linked a few specific postures with emotional experiences, for example backward ear posture, which is associated with unfamiliar and uncontrollable unpleasant situations, could express fear \cite{boissy2011cognitive}. 
Identifying the pain or suffering of animal (like sheep) is an essential aspect of animal welfare and is very helpful to both researchers and farmers. As an example shown in Fig.\ref{fig:sheepinpain}, the sheep on the right is suffering heavily from pain while the sheep on the left is in a normal condition. Experts on animal welfare research are able to pick up several distinguishable patterns of sheep-in-pain such as orbital tightening, abnormal ear position and abnormal nostril and philtrum shape. In order to identify those features automatically, localising the corresponding landmarks on sheep face is very essential, which is conceptually very similar to human facial landmarks localisation (also called face alignment). As a classical problem in computer vision, face alignment has been intensively studied in the past decades due to its wide applications for example face recognition, facial expression recognition, avatar animation, etc. Several recent methods such as \cite{suncvpr2012,burgos2013robust,kazemi2014one,renface,yang2015robust,xiong2013supervised,zhang2014facial} have reported close-to-human performance on the academic databases such as LFW \cite{LFWTech}, LFPW \cite{belhumeur2011localizing} and HELEN \cite{interactiveECCV2012}. 

However, we meet several obstacles when we apply the state of the art algorithms directly to real data, for both human and sheep facial landmark localisation. First, unlike the benchmark dataset for human face alignment, in which a large number of landmarks are often annotated, the number of facial landmarks in practice is usually smaller, due to the annotation cost and fewer landmarks of interest. Second, both human face and sheep face show big head pose variations in real world given the uncontrollability. It usually results in localisation failures.

In this paper we deal with the problems mentioned above. We build our localisation algorithm on top of the Cascaded Pose Regression (CPR) framework, given its good performance in facial landmarks localisation in the wild. There has been a series of works with incremental improvement one after the other including \cite{burgos2013robust,suncvpr2012,dollar2010cascaded}. The most recent work RCPR (\cite{burgos2013robust}) introduced interpolated shape-indexed features used in each regression. It demonstrated better robustness against large pose variations and shape deformations, compared to the closest landmark indexed feature in \cite{suncvpr2012}. However, the two-point-interpolation method limits the feature extraction space, especially when the number of facial landmarks is small. Landmark sparsity is often the case when we need to annotate a new training dataset given a limited amount of time or only a small number of landmarks needed. To overcome those issues we make the following contributions. 

\begin{itemize}
\item We propose a new feature extraction scheme, called triplet-interpolation feature (TIF) for cascaded pose regression. It uses three anchor landmarks to calculate a shape-indexed feature. It is more robust to large head pose variation and shape deformation. More importantly, with this scheme, features can be extracted from the facial area with no restriction. 
\item We propose an augmentation scheme for training sample  to deal with the issue of imbalanced training data distribution. This scheme sets the augmentation number of each training sample to be negatively correlated to its value in the probability density function of the training data. More intuitively, we augment the minority training samples with more random initialisations and vice versa. 
\end{itemize}

We have carried out experiments on both human and sheep facial landmarks localisation and demonstrate the benefits of our proposed methods under the situation of sparse landmarks and large head pose variations. It also shows competitive overall performance comparing to other related methods.  

The reminder of the paper is organized as follows. In section \ref{sec::relatedwork} we present related work. Then we introduce the triplet-interpolation features and the augmentation scheme in section \ref{sec::method}. In section \ref{sec::exp} we evaluate our proposed methods on both human and sheep facial landmarks localisation and in section \ref{sec::conlusion} we draw some useful conclusions. 

\section{Related work\label{sec::relatedwork}}
\subsection{Facial landmarks localisation}
Facial landmarks localisation has made considerable progress in recent years and a large number of methods have been proposed. Two types of source information are usually used: facial appearance and shape information. Based on whether a method has an explicit detection model for an individual landmark or not, we categorise them into local-based methods and holistic-based methods. The methods in the former category usually rely on explicit discriminative local detection and usually use deformable shape models to regularise the local outputs while the methods in the latter category directly regress the shape (the representation of the facial landmark locations) in a holistic way.

Local based methods usually consist of two parts: local experts and spatial shape models. The former describes how image around each facial landmark looks like in terms of local intensity or colour patterns while the latter describes how face shape varies. There are  three main types of local feature detection. (1) Classification methods include Support Vector Machine (SVM) classifier \cite{rapp2011multiple,belhumeur2011localizing} based on various image features such as Gabor \cite{vukadinovic2005fully}, SIFT \cite{lowe2004distinctive},  Discriminative Response Map Fitting (\textbf{DRMF}) by dictionary learning \cite{asthana2013robust} and multichannel correlation filter responses \cite{Kiani_2013_ICCV}. (2) Regression-based approaches include Support Vector Regressors (SVRs)\cite{martinez2012local} with a probabilistic MRF-based shape model, Continuous Conditional Neural Fields (\textbf{CCNF})\cite{baltruvsaitis2014continuous}. (3) Voting-based approaches are also introduced in recent years, including regression forests based voting methods \cite{cootesECCV2012,dantone2012real,yangiccv2013} and exemplar based voting methods \cite{smithnonparametric,shen2013detecting}. One typical shape model is the Constrained Local Model (CLM) \cite{cristinacce2006feature}. There are some other shape models such as RANSAC in \cite{belhumeur2011localizing}, graph-matching in \cite{Zhou_2013_ICCV}, Gaussian Newton Deformable Part Model (\textbf{GNDPM}) \cite{tzimiropoulos2014gauss} and mixture of trees \cite{devacvpr2012face}.

\begin{table*}[!hbtp]
\centering
\caption{Holistic methods and their properties.}
\label{tab::holisticmethods}
\begin{tabular}{lllllll}
\hline
Methods          & \textbf{RCPR} \cite{burgos2013robust}         &  \textbf{ESR} \cite{asthanaincremental} & \textbf{LBF}  \cite{renface}        & \textbf{TREES} \cite{kazemi2014one}        & \textbf{SDM} \cite{xiong2013supervised}     & \textbf{TCDCN} \cite{zhang2014facial} \\
features         & pixel diff.     &pixel diff.    & forest on pixels         & pixel   & SIFT    & ConvNet feature\\
regressor         & random ferns & random ferns & linear & random trees & linear &   ConvNet    \\
\hline
\end{tabular}
\end{table*}
Holistic methods have gained higher popularity in recent years. Most of them work in a cascaded way similar to the classical Active Appearance Model (AAM) \cite{cootes2001active}. 
We list very recent holistic methods as well as their properties in Table~\ref{tab::holisticmethods}.  These methods work in a similar cascaded framework but differ from each other mainly in three aspects. First, how to set up the initialisations; Second, how to calculate the shape-indexed features; Third, what type of regressor is applied at each iteration. Feature extraction and regression are usually interdependent. As can be seen, several methods have investigated using simple pixel difference (diff.) features that is calculated from the current shape. Random ferns and random trees are widely used for regression. Using raw pixel difference feature makes the algorithm very efficient. In our testing, the method ESR, RCPR, LBF and TREES with c++ implementation process a standard face image in mini-seconds on an i7 desktop with a single core. This is a great advantage in systems that are designed to process a large number of faces, for example to analyse a group of sheep at the same time. SDM has been widely applied given its good performance of the publicly available model. It runs at around 30 frames per second. TCDCN has applied deep learning approach for face alignment by multi-task learning, but training such a model usually requires a big dataset with multiple additional annotations such as head pose, w/o glasses, etc. 

There are several other approaches for holistic face alignment such as occlusion detection based methods by \cite{fowlkesocclusion,yang2015robust}, combined local and holistic method in \cite{alabort2015unifying}, SDM variants including the global SDM \cite{xiong2015global} and shape searching in \cite{zhu2015face}. Due to their different setting and limited space, we will not compare them in our experiments. 

\subsection{Data imbalance}
The data imbalance problem is of particular importance in real world scenarios as the available data usually follows a long tail distribution. Data imbalance has been widely studied in classification problems, i.e., a few classes are abundant while others only have a limited number of samples \cite{he2009learning}. State of the art solutions include sampling methods (e.g. under-sampling \cite{liu2009exploratory} and SMOTE over-sampling \cite{chawla2002smote}), cost-sensitive learning \cite{elkan2001foundations, he2009learning}. On the contrary, very little attention has been paid on data imbalance in regression problem (like our facial landmark localisation). This is mainly due to the fact that the data imbalance is difficult to be noticed given the continuity and the usually high dimensionality of the output space. Thus in this paper we investigate how to adapt the approach of tackling class imbalance to regression problem.

\section{Method\label{sec::method}}

In this section, we first briefly review the general cascaded pose regression (CPR) approach, on which our localisation algorithm has been built. Then we introduce the triplet-interpolated features. Following that, inverse proportional augmentation is discussed in details as an approach to deal with imbalanced training data.

\subsection{General CPR and RCPR} 

The shape of a human or sheep face is represented as a vector of landmark locations, i.e., $S=(\mathrm{y}_1,...,\mathrm{y}_k,...,\mathrm{y}_K) \in\mathbf{R}^{2K}$, where $K$ is the number of landmarks. $\mathrm{y}_k \in \mathbf{R}^2$ is the 2D coordinates of the $k$-th landmark. CPR is formed by a cascade of $T$ regressors, $R^{1...T}$. Shape estimation starts from an initial shape $S^0$ and progressively refines the pose. Each regressor refines the pose by producing an update, $\Delta S$, which is added up to the current shape estimate, that is,
\begin{equation}
S^t = S^{t-1} + \Delta S.
\end{equation}
The update $\Delta S$ is returned by the regressor that takes the previous pose estimation and the image feature $I$ as inputs:
\begin{equation}
\Delta S = R^t(S^{t-1},I)
\end{equation}
The CPR is summarized in Algorithm \ref{alg::algorithm1} \cite{dollar2010cascaded}.This CPR framework differs from the classic boosted approaches mainly in the feature re-sampling process. More specifically, instead of using the fixed features, the input feature for regressor $R^t$ is calculated relative to the current pose estimation, thus in turn introduces geometric invariance into the cascade process and shows good performance in practice. This is often referred as pose-indexed features as in \cite{dollar2010cascaded}. The idea of sampling features from current pose estimation is later used in \cite{suncvpr2012,kazemi2014one}. To strengthen the geometric invariance, instead of extracting features from the closest landmarks, RCPR \cite{burgos2013robust} utilizes a different feature-indexing method ($h^t(I,S^{t-1})$), namely the interpolated shape-indexed features. The features are extracted with reference to two shape points. \cite{burgos2013robust} has proven that RCPR is more robust to large pose variations than the general CPR. 
\begin{algorithm}
\caption{Cascaded Pose (shape) Regression}
\label{alg::algorithm1}
\begin{algorithmic}[1]
    \Require{Image $I$, initial pose $S^0$} 
    \Ensure{Estimated pose $S^T$}
    \For {$t$=1 to $T$}
    \State $f^t = h^t(I,S^{t-1})$\Comment{Shape-indexed features} 
    \State $\Delta S = R^t(f^t)$\Comment{Apply regressor $R^t$}
    \State $S^t = S^{t-1}+\Delta S$\Comment{update shape}
     \EndFor
\end{algorithmic}
\end{algorithm}



\subsection{Triplet-Interpolated Feature (TIF)}
The above CPR scheme and its variants are very popular given its high computational efficiency and localisation accuracy. In each iteration, random ferns or random forests takes raw pixel values as input features, which in turn become essential to fast convergence in the cascaded learning. Prevalent pixel-indexing features intend to be invariant with respect to pose variation. That is to say, the indexed pixels referencing to same shape points are expected to have same semantic meaning across different samples. Such efforts have been made in \cite{suncvpr2012}, which applied shape-indexed features, and in \cite{burgos2013robust}, which achieved stronger geometric invariance with the interpolated shape-indexed features.

\begin{figure}[!t]
 \centering
\includegraphics[width=0.4\textwidth,height=0.4\textwidth]{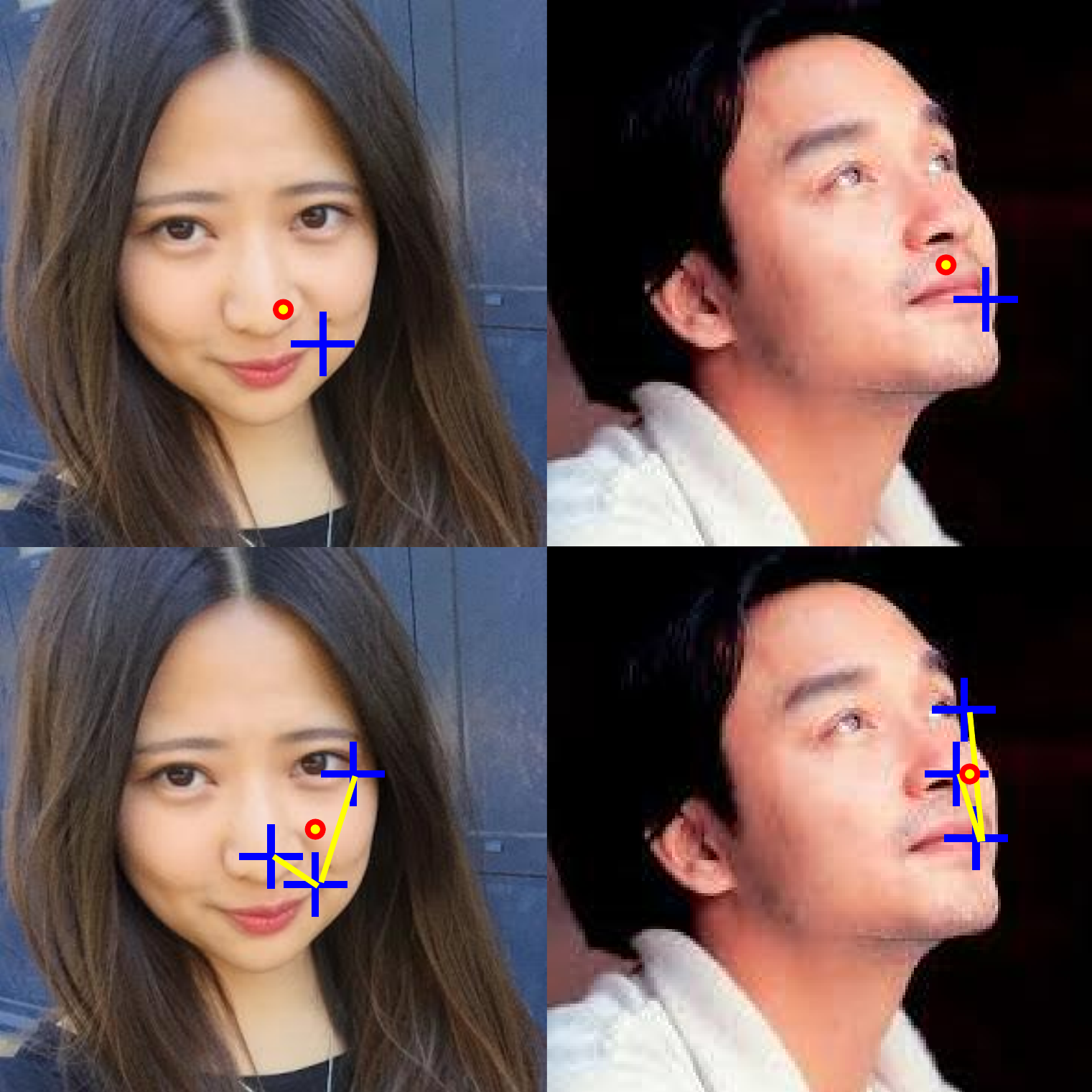}
\caption{Pixels indexed by the same local coordinates should have the same semantic meaning. The triplet-interpolated feature shows its feature invariance to large pose variation in the right bottom figure.}
\label{fig::tifvsesr}
\end{figure}

\begin{figure}[!t]
 \centering
\subfloat[]{\includegraphics[width=0.20\textwidth,height=0.20\textwidth]{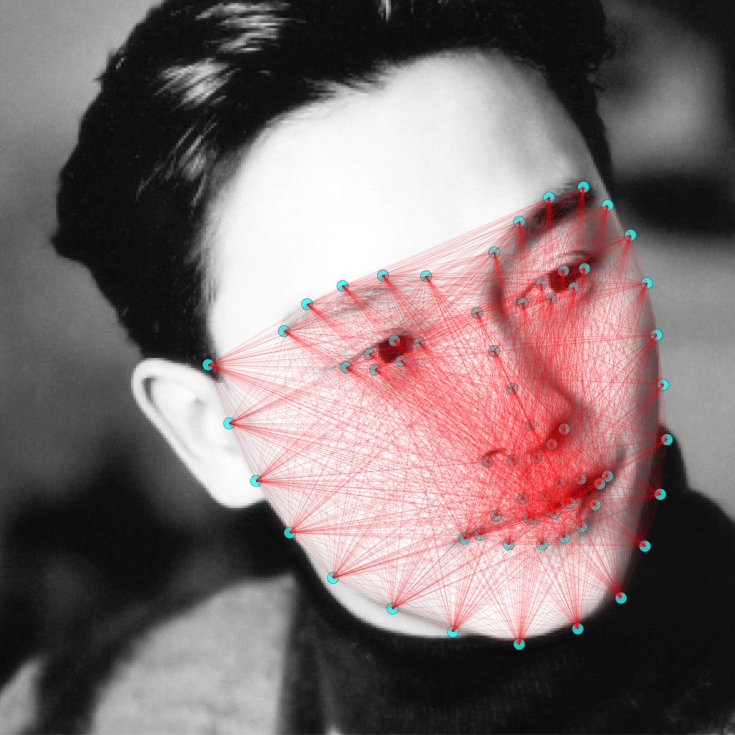}\label{fig::line_shape_whole}}
\subfloat[]{\includegraphics[width=0.20\textwidth,height=0.20\textwidth]{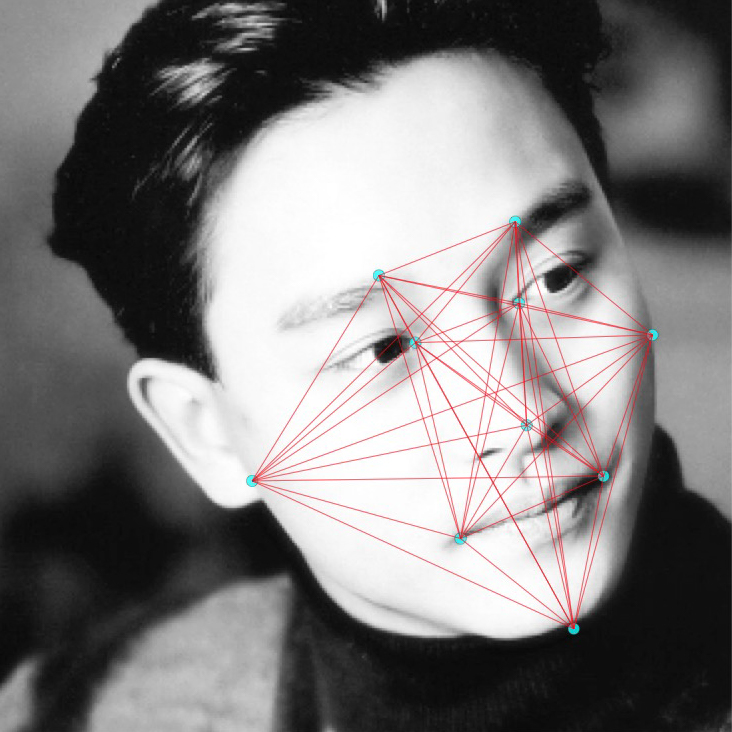}\label{line_shape}}\\
\subfloat[]{\includegraphics[width=0.20\textwidth,height=0.20\textwidth]{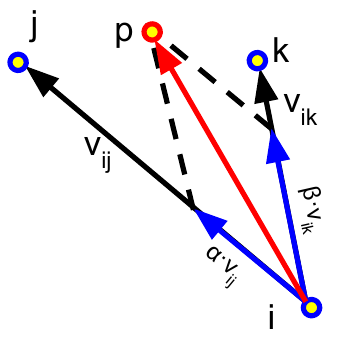}\label{fig::vector_graph}}
\subfloat[]{\includegraphics[width=0.20\textwidth,height=0.20\textwidth]{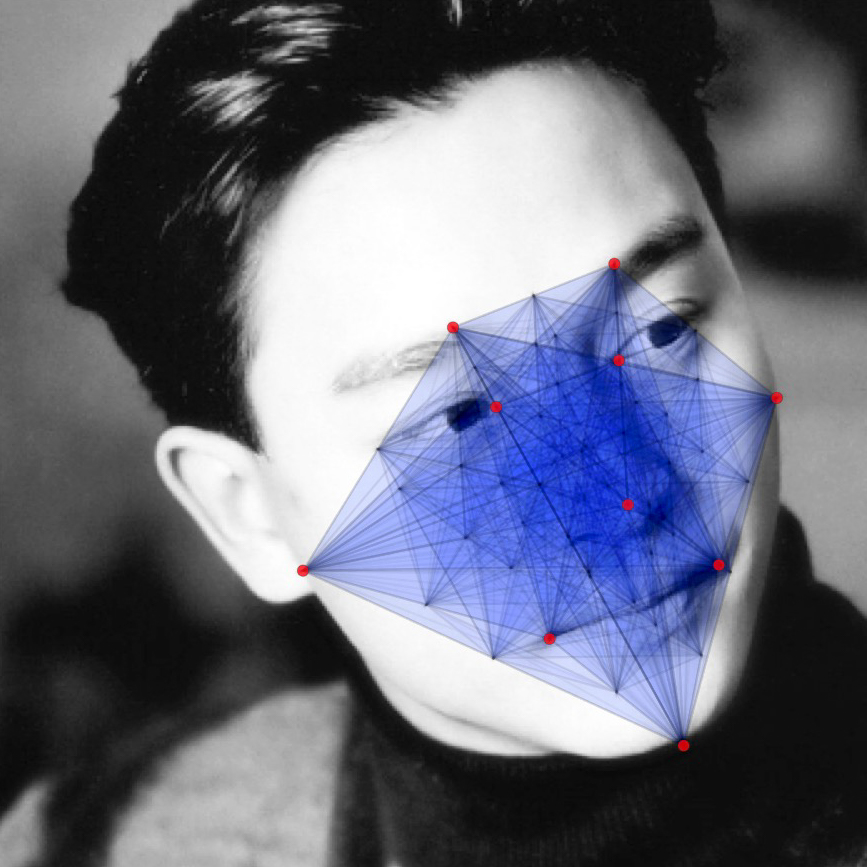}\label{fig::tri_shape}}\\
\caption{The red lines in (a)(b) show the available area for feature extraction when we use the linear-interpolated shape-index features. (c) and (d) illustrate the concept of our Triplet-interpolated features and its available feature region. (b)(d) together show that how the new indexing method extends the available area for feature extraction when the shape annotation is sparse. }
\label{fig::comparison}
\end{figure}

However, the interpolated shape-indexed features in RCPR has a fundamental drawback. It can only draw features that are lying on the line segment between two landmarks. As example shows in Fig ~\ref{fig::line_shape_whole}, features can be extracted from a rich area of the face when the landmarks are dense. However it becomes problematic when the facial landmarks are sparse. Features can only be extracted from very restricted locations (see Fig. \ref{line_shape}). This limits the randomness of feature extraction. 

To combine the benefits of geometric invariance and avoid its limitations, we propose a new indexing approach, namely Triplet-interpolated feature(TIF), as shown in \ref{fig::vector_graph}. The indexing process works in the following way: Out of every group of three randomly selected landmarks, one is randomly chosen and assigned as the primary point. Then two vectors, from the primary to the rest two, can span the whole plane by linear combination. By setting the parameters of the linear combination, a position can be selected within the spanned area, as shown in Fig ~\ref{fig::vector_graph}. The location of the point $p$ indexed by TIF is represented as:
\begin{equation}
\mathrm{p} (S,i,j,k,\alpha, \beta) = \mathrm{y}_i + (\alpha \cdot \vec{v}_{ij} + \beta \cdot \vec{v}_{ik})   
\label{equ::vector}
\end{equation}
where $S$ is the current shape and $i$, $j$, $k$ are landmark indexes. $\vec{v}_{ij} = \mathrm{y}_j - \mathrm{y}_i$ is the vector from the position of $i$ to the position of $j$. $\alpha$ and $\beta$ are the random ratios that control the position of the indexed point. Compared to the original closest landmark indexed feature in \cite{suncvpr2012}, the TIF has two main advantages: 1) it is computationally cheaper since it does not have the shape transformation step; 2) it is more robust to large head pose variation given the triplet interpolation property, as shown in Fig.~\ref{fig::tifvsesr}. Compared to the two-point-interpolated-feature in RCPR \cite{burgos2013robust}, it is able to extract features from a much wider range, especially when the landmarks are sparse. We will show the benefits of using TIF in the experiment section. Apart from the feature extraction process, we follow the cascaded pose regression process used by ESR \cite{suncvpr2012} and RCPR \cite{burgos2013robust}. Note that in this paper, we only use the feature extraction part of RCPR as the occlusion estimation part requires landmark-wise occlusion annotation. In this way we also make the benefits of feature extraction clearer.  



\subsection{Negatively Correlated Augmentation (NCA)}

Before introducing our data augmentation scheme, we first analyse the data distribution of the benchmark database for human facial landmark localisation, i.e., 300w, which is a benchmark database for human facial landmark localisation. It consists of face images from  AFW \cite{devacvpr2012face}, HELEN \cite{tan2009enhanced}, LFPW \cite{belhumeur2011localizing} and the newly annotated iBug \cite{sagonas300}. We partition it to 3148 training images and 689 test images. Training images are from AFW (337 images), HELEN training set (2000 images) and LFPW training set (811 images), and test images are from HELEN test set (330 images), LFPW test set (224 images) and iBug (135 images). 

Because it is impractical to analyse the data distribution directly on the output space given its high dimensionality, we ignore individual face difference and small facial deformation. Then facial landmarks distribution is mainly affected by head pose variations, which lie in low dimensional manifold. Therefore, we analyse the distribution of head poses. Since head pose is not provided by the database, estimated head pose information for each face is derived from the annotated facial landmarks. To this end, we fit a mean 3D model (68 facial points) of a head to the annotated points in
the image. Then we feed the set of corresponding 3D and 2D points to the POSIT \cite{dementhon1992model} algorithm which produces the head pose information. 

\begin{figure}[!t]
 \centering
\subfloat[]{\includegraphics[width=0.24\textwidth,height=0.18\textwidth]{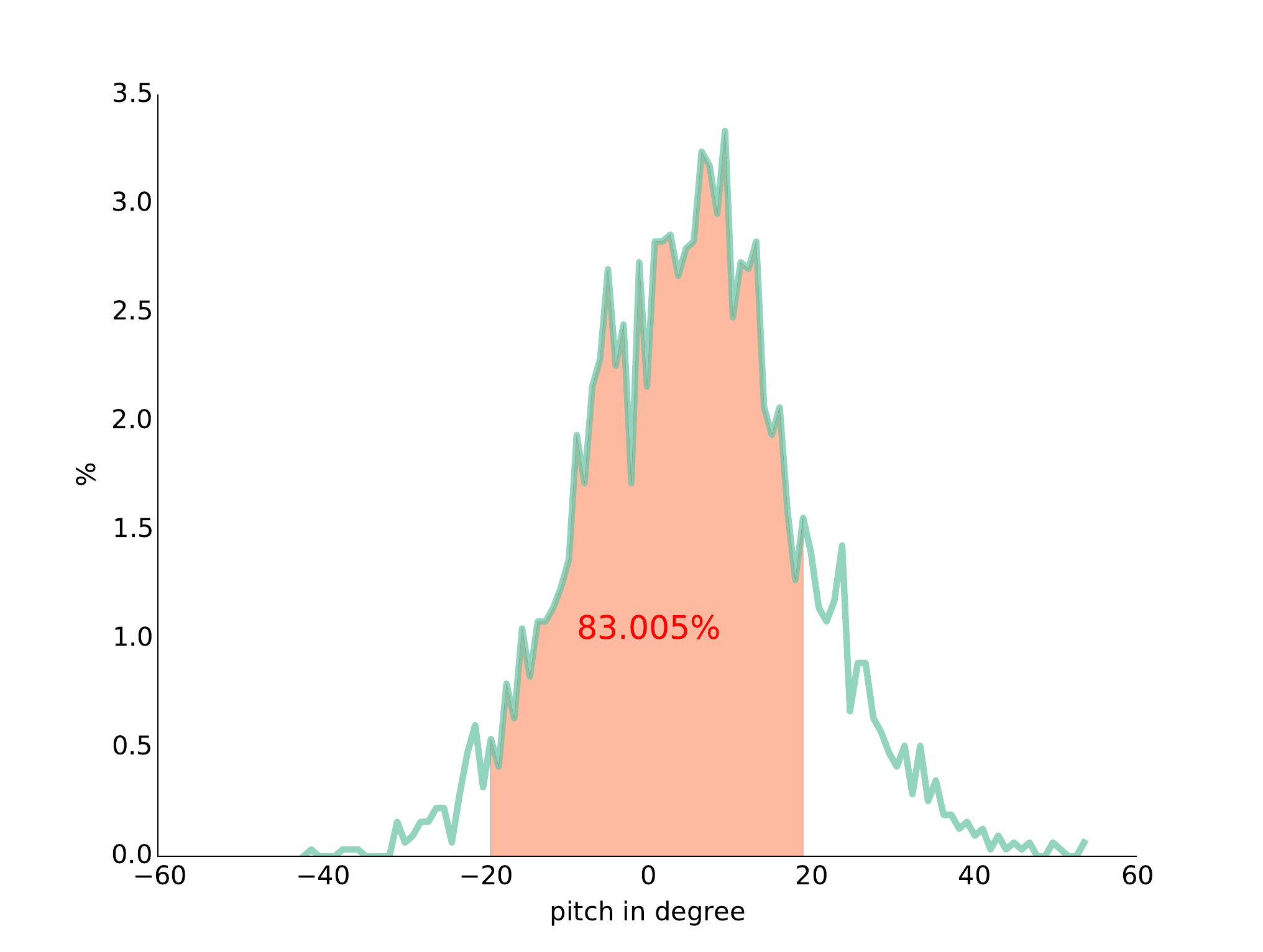}\label{fig::headposepitch}}
\subfloat[]{\includegraphics[width=0.24\textwidth,height=0.18\textwidth]{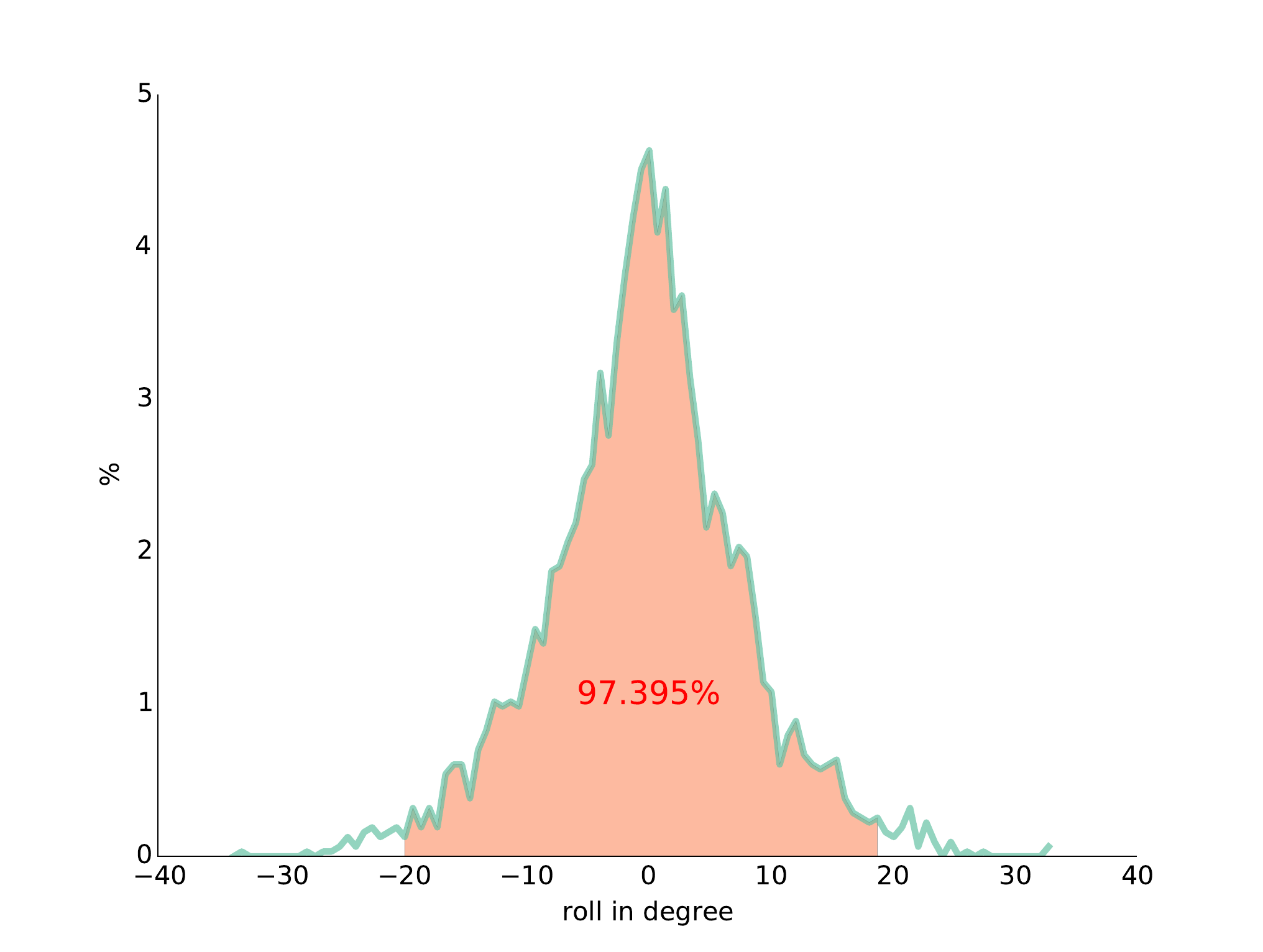}\label{fig::headposeroll}}\\
\subfloat[]{\includegraphics[width=0.24\textwidth,height=0.18\textwidth]{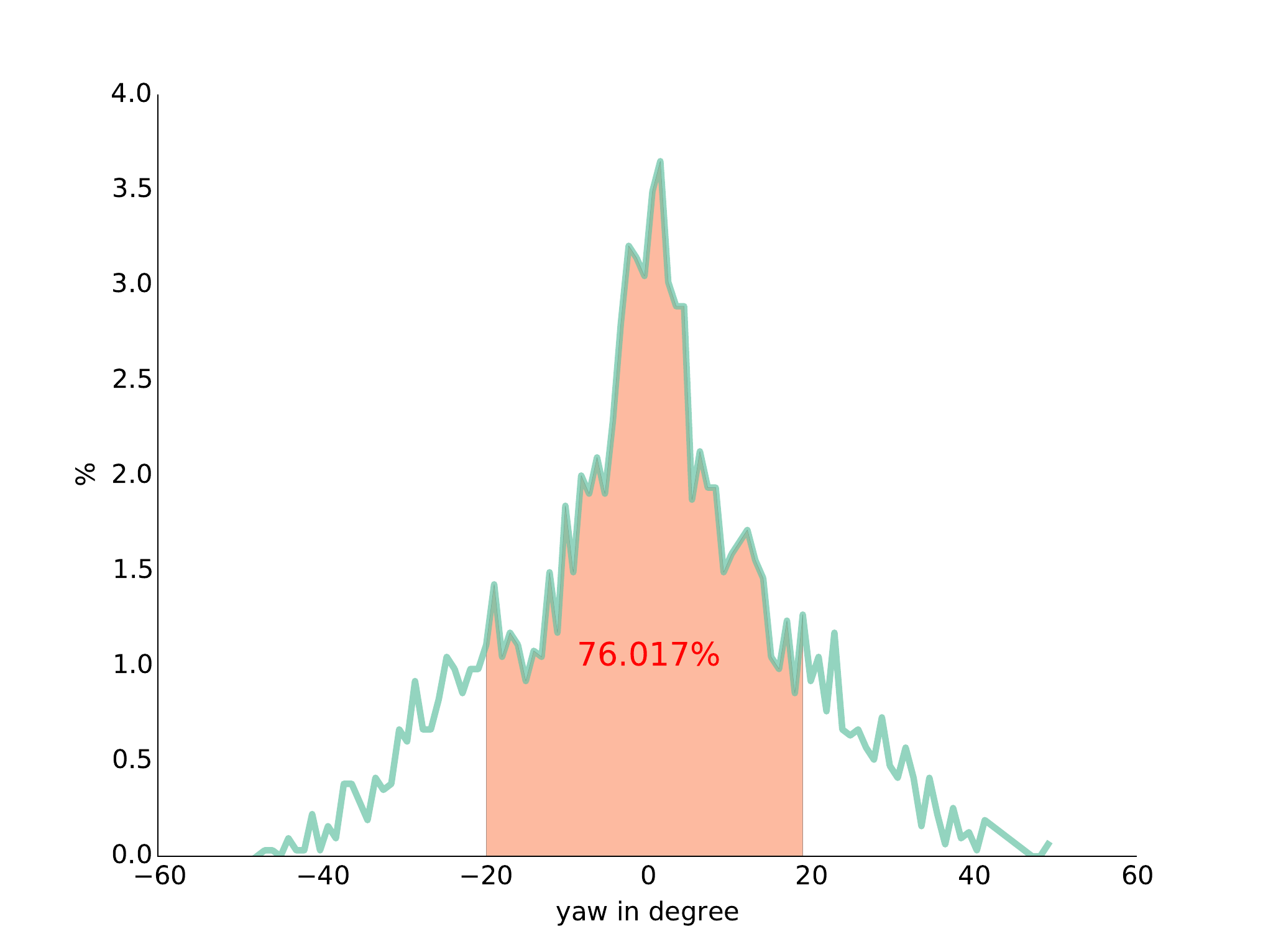}\label{fig::headposeyaw}}
\subfloat[]{\includegraphics[width=0.24\textwidth,height=0.18\textwidth]{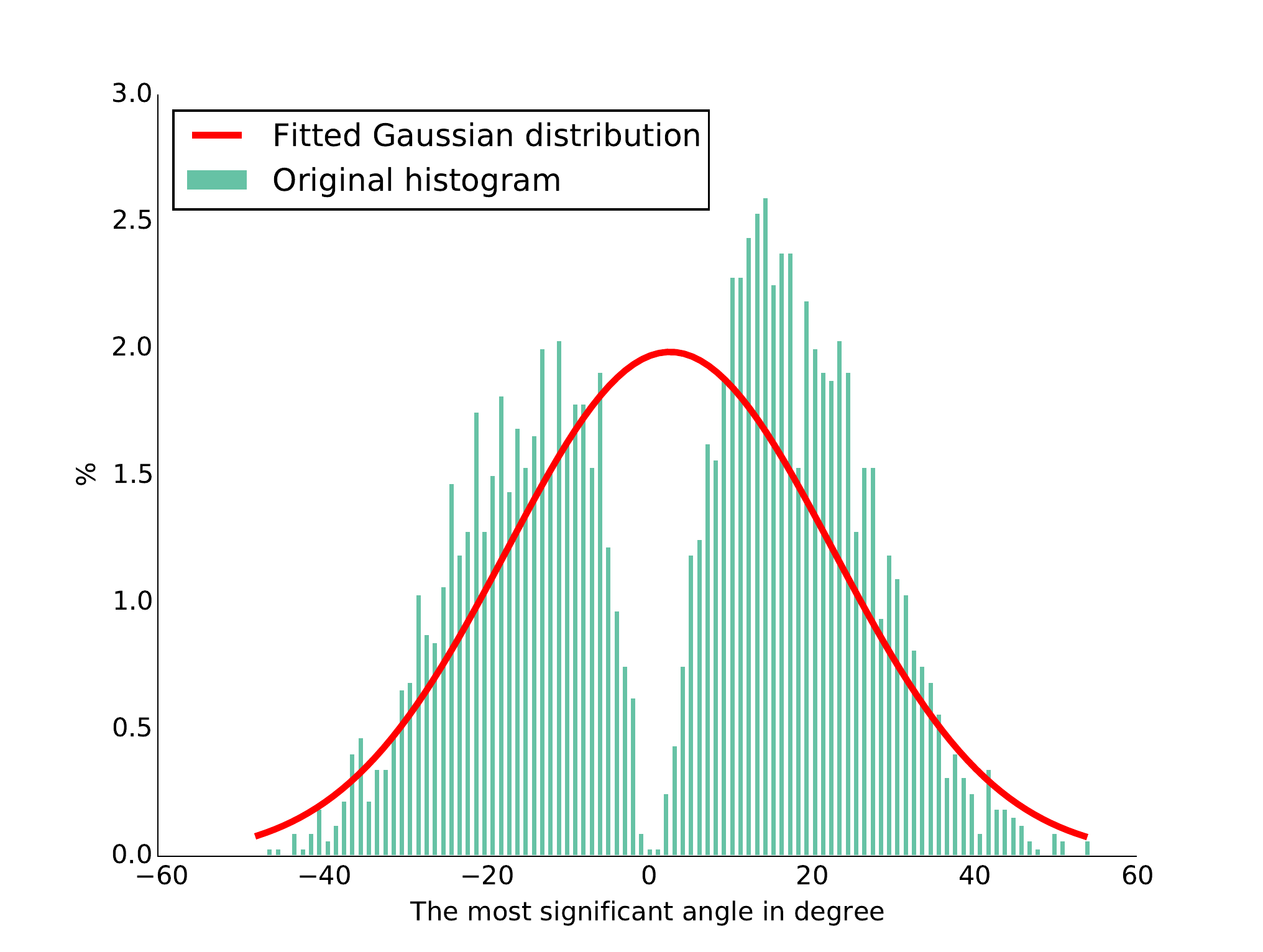}\label{fig::headposegaussian}}\\
\subfloat[]{\includegraphics[width=0.5\textwidth,height=0.24\textwidth]{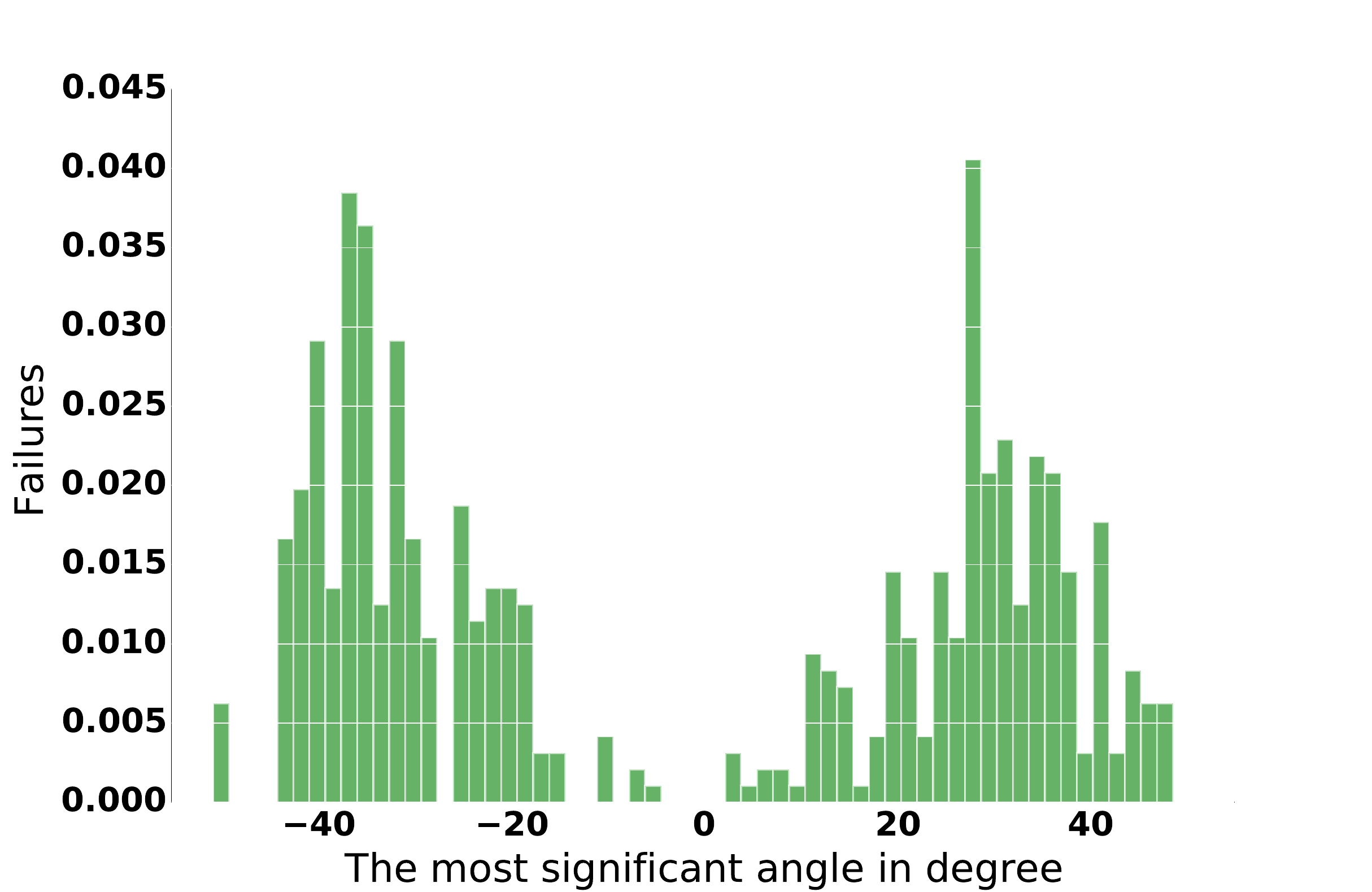}\label{fig::failure_dis}}
\caption{(a) (b) (c) show the histogram of the head pose pitch, roll and yaw angles respectively in 300w training set. (d) shows the fitted Gaussian curve and the histogram of the most significant angle of the training samples. (e) is the histogram of failures from several state of the art models trained on 300W (ESR, TREES, RCPR and SDM). }
\label{fig::headposedis}
\end{figure}

As shown in Fig.~\ref{fig::headposedis}, the majority of training samples distribute near frontal angles. More than 97\% of the samples lie within roll angle range between -20$^{\circ}$ and 20$^{\circ}$. For pitch and yaw angle, such percentages are 83\% and 76\% respectively. For each training sample, we calculate the most significant rotation angle, i.e., the angle with the biggest absolute value. Then we fit a Gaussian curve on all the training samples as shown in Fig. \ref{fig::headposegaussian}. 

We ran several models on the test images including the Explicit Shape Regression (ESR) \cite{suncvpr2012}, the Robust Cascaded Pose Regression (RCPR) \cite{burgos2013robust}, the Supervised Descent Method (SDM) \cite{xiong2013supervised}, and the TCDCN \cite{zhang2014facial}. Then we recorded their failures, i.e. a sample with mean localisation error bigger than 0.1 inter-ocular-distance (IOD). The overall distribution is shown in Fig.~\ref{fig::failure_dis}. Despite these methods being modelled in very different ways, their failures are quite similar. Only a few failures are within angle range between -20$^{\circ}$ and 20$^{\circ}$, where the majority of training samples distribute.  To this end, we can conclude that the imbalanced distribution of training data has heavy impact on testing performance, regardless of the algorithm design. 

In the framework of cascaded shape regression, data augmentation is usually carried out during training time. More specifically, for one face image sample, several initialisations are generated by Monte Carlo method. This procedure has been used in ESR, RCRP and SDM and the augmentation number is usually fixed. We propose a simple augmentation scheme, under which the amount of augmentation of each training sample is negatively correlated to the value on the fitted Gaussian curve (Fig.~\ref{fig::headposegaussian}). Conceptually, this is similar to over-sampling in classification problem but each augmented sample becomes unique in our case because of the initialisation difference. More specifically, the augmentation number $m_{x}$ of training sample $x$ is calculated as:
\begin{equation}
m_{x} = a \cdot \mathcal{N}(x_{\text{pose}}) + b
\end{equation}
where $x_\text{pose}$ the head pose of $x$, $\mathcal{N}(\cdot)$ the fitted Gaussian distribution. $a$ is a negative variable that controls the slope and $b$ is a bias term that controls the bounds of augmentation numbers. We use two pairs of values (the maximum and the minimum) to fit this linear equation with a constrain that the total number after augmentation is equal to the baseline augmentation scheme.

\section{Evaluation\label{sec::exp}}
\subsection{Sheep face experiment}
\begin{figure*}[!t]
 \centering
\includegraphics[trim={2cm 7cm 2cm 2cm},clip, width=0.99\textwidth,height=0.42\textwidth]{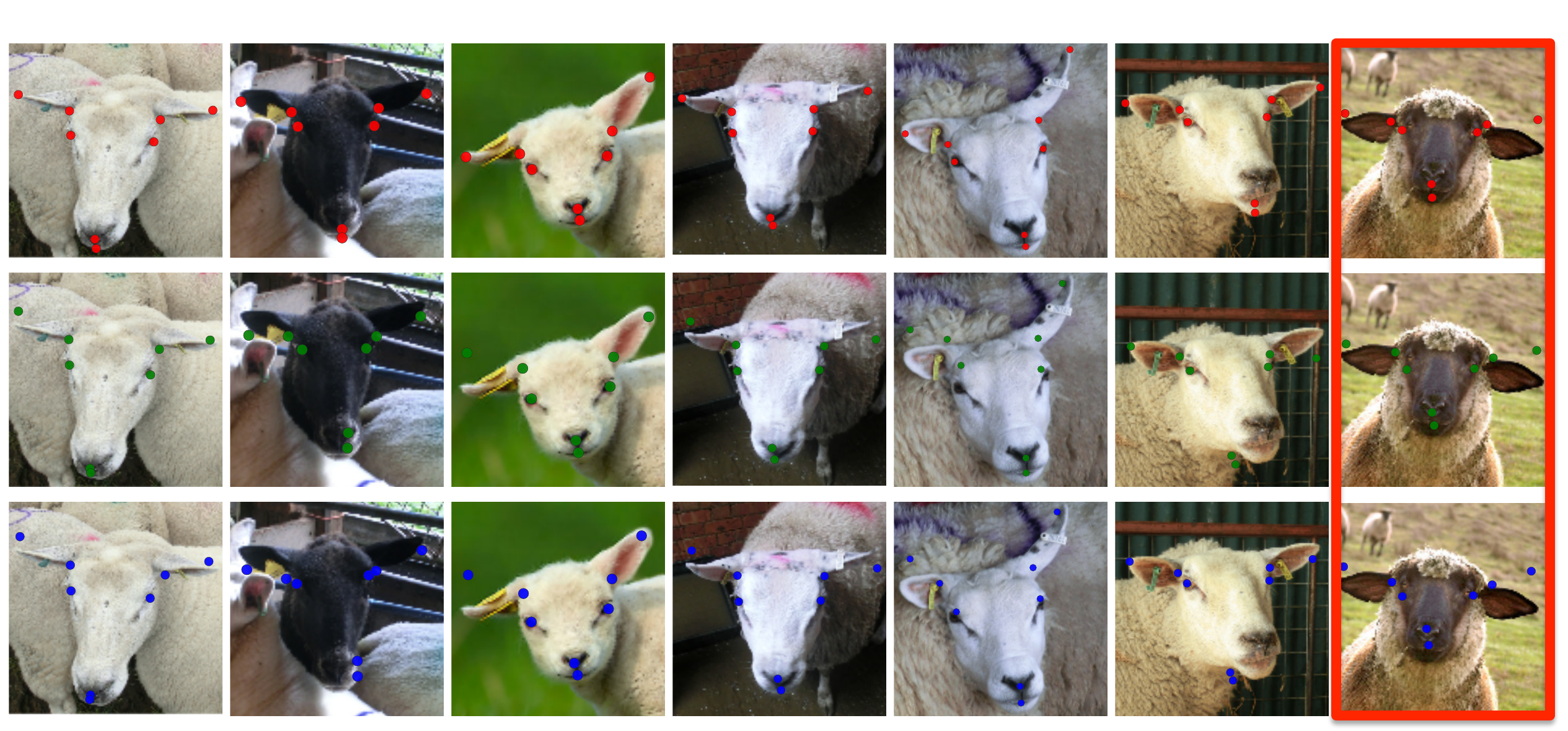}
\caption{Landmarks localisation on example sheep face images. From top to bottom show the result of our TIF method, RCPR and ESR respectively. The final column shows a failure example of our method.}
\label{fig:comparesheepfaces}
\end{figure*}

We collected 600 sheep face photos from an animal research centre. We manually labelled the bounding boxes and 8 landmarks on faces as shown in Fig.\ref{fig:sheepinpain}. We trained a structured SVM sheep face detector based on HOG features using  dlib  \cite{dlib09}. Using a few hundred sheep face images is sufficient to train a sheep face detector which can be used in real videos. In our sheep facial landmarks localisation, as usual, we assumed the face bounding boxes are available. We randomly split the 600 sheep faces into a training set (500) and a testing set (100). Then we trained our TIF model, ESR and RCPR using the same training set. We set the augmentation number to 20 for all these methods. We repeated this random process for 5 times, and recorded all the results. Since our test set is not big, we directly report the sorted sample-wise mean error (normalised by sheep face size) of the 100 images. For each index, the value is the average over 5 runs. As can be seen, on a small dataset with sparse landmarks, our method outperforms the baseline methods by a large margin. Around 90\% of the sheep images are localised with mean error less than 10\% of the face size. Some example images with comparison to other methods are shown in Fig.~\ref{fig:comparesheepfaces}. The sheep face image in our collected dataset exhibits a wide range of diversity: sheep breed, facial colour, lighting condition, background, occlusion, head pose, ear posture, etc. 

\begin{figure}[!t]
 \centering
\includegraphics[width=0.45\textwidth,height=0.27\textwidth]{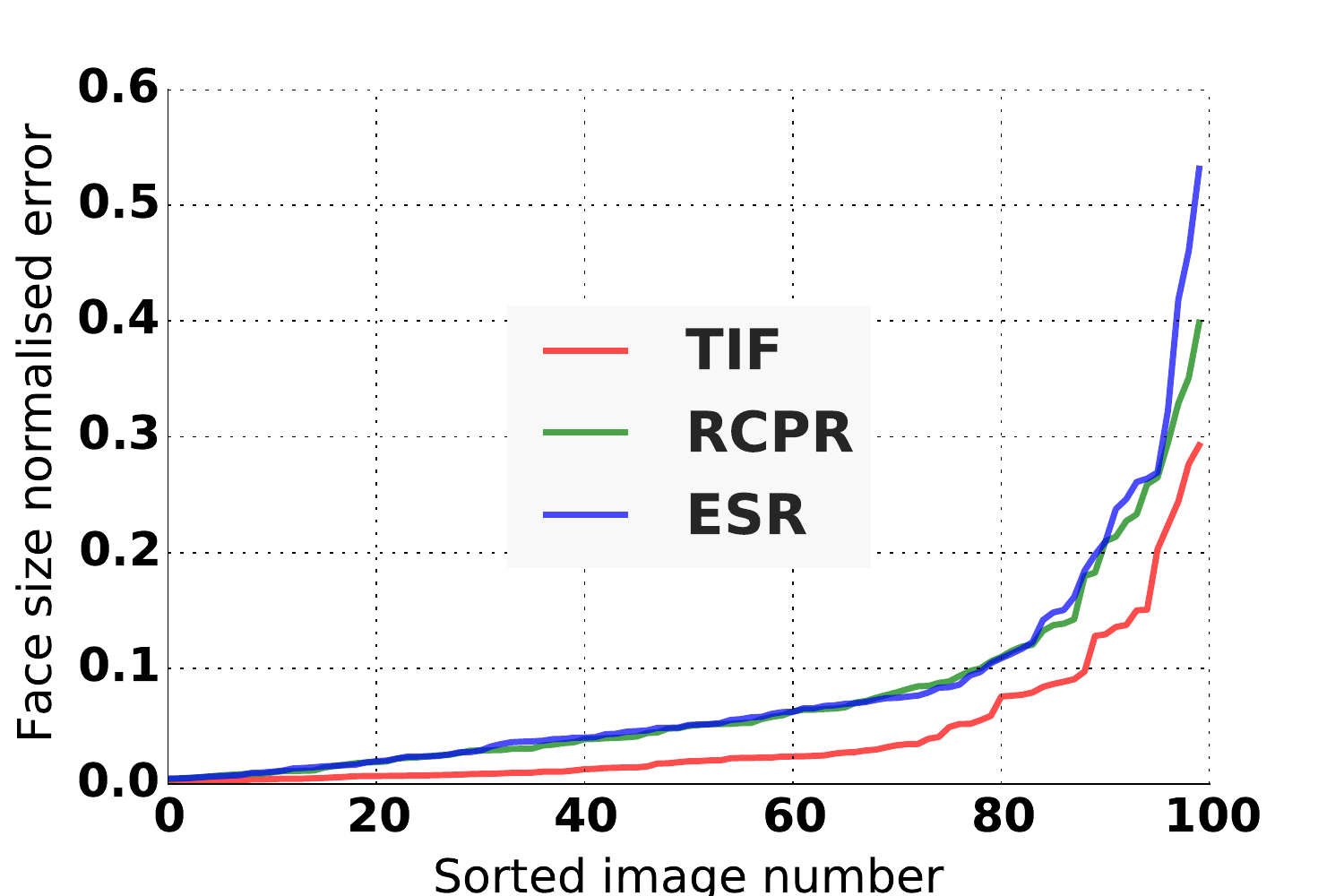}
\caption{Performance comparison on our sheep face dataset. (Lower is better.)}
\label{fig:comparesheep}
\end{figure}

\subsection{Human face experiment}
In order to further evaluate the proposed schemes, we carry out experiments on human face alignment benchmark database, i.e., 300w. Recall that we split the publicly available database into training set (3148 images) and testing set (689 images). We have implemented and trained the baseline models (ESR and RCPR) on the same training images. Note that when implementing the RCPR algorithm we only used their method of feature indexing (interpolation by two landmarks) but not their occlusion modelling since there is no occlusion annotation for training. Thus for ESR, RCPR and our TIF method, the only difference is their feature extraction step. During testing time, we also initialised them with the same random shapes for a fair comparison. 
\begin{figure}[!t]
 \subfloat[]{\includegraphics[width=0.25\textwidth,height=0.22\textwidth]{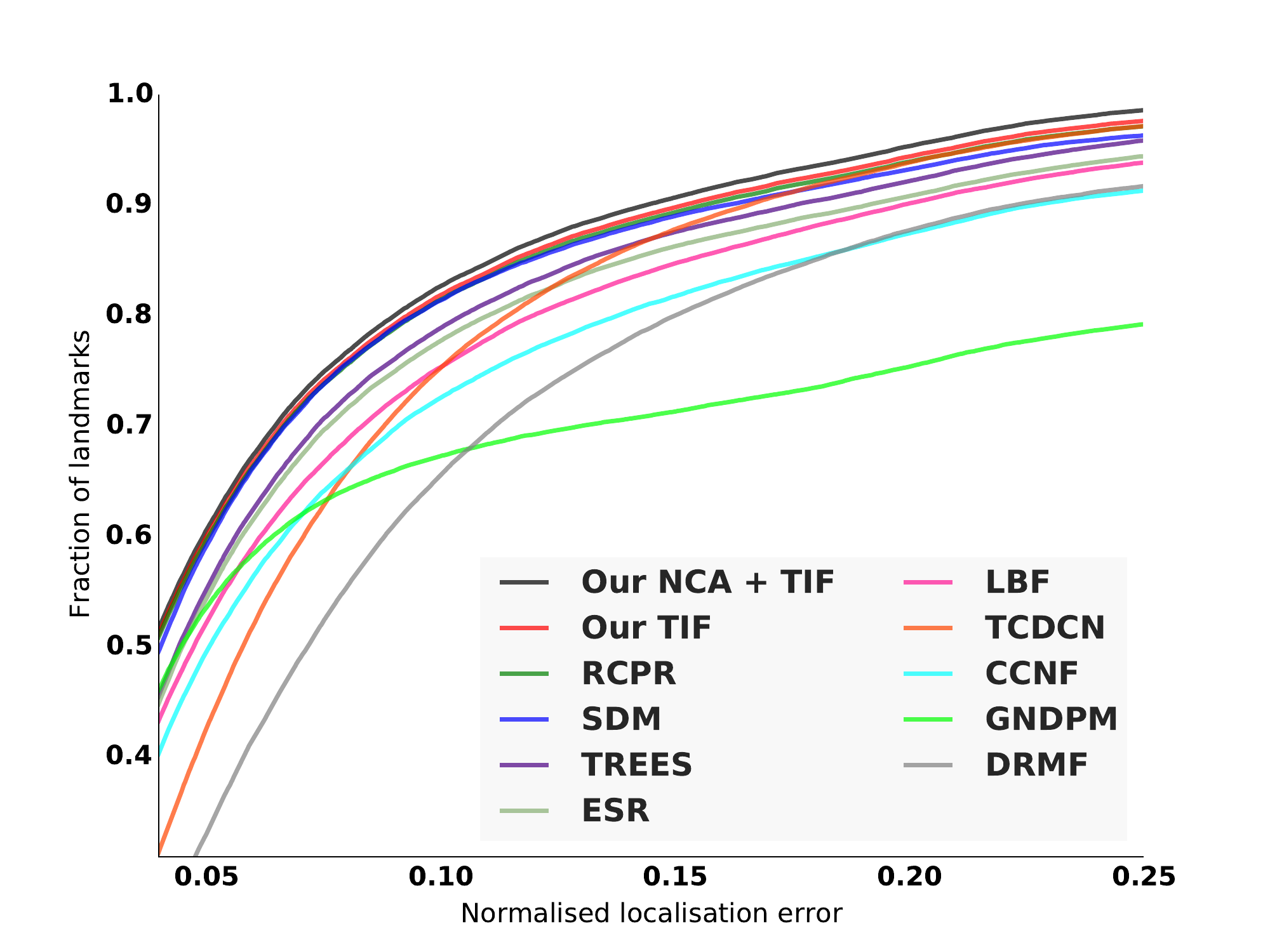}\label{fig::compare68p}}
\subfloat[]{\includegraphics[width=0.25\textwidth,height=0.22\textwidth]{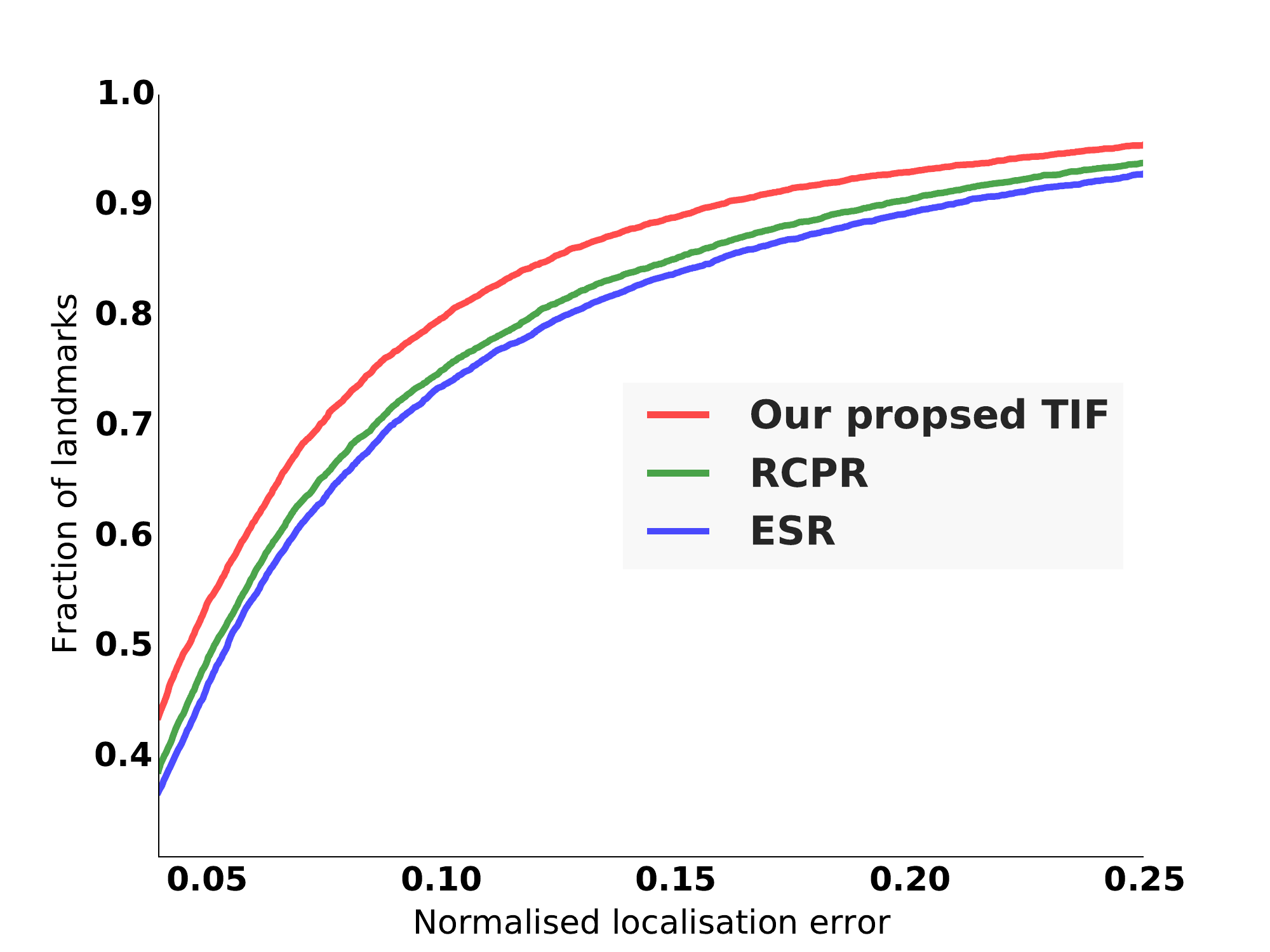}\label{fig::compare10p}}
\caption{Performance comparison on dense (left) and sparse (right) facial landmarks. (Higher is better.)}
\label{fig:compare68pand10p}
\end{figure}
\begin{figure}[!t]
\centering
\includegraphics[trim={1cm 12cm 1cm 0},clip, width=0.45\textwidth,height=0.24\textwidth]{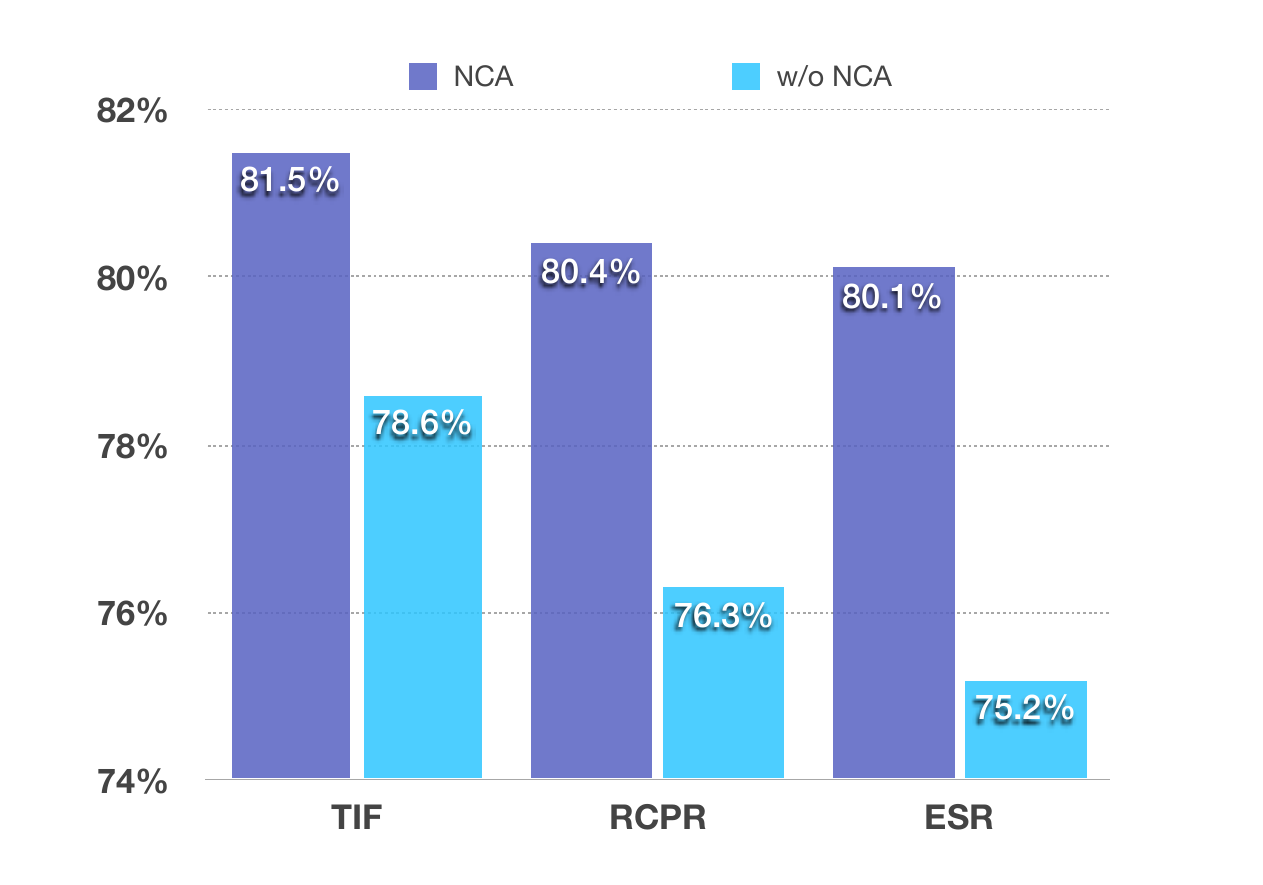}
\caption{Successful-localisation-rate comparison for methods with and w/o our proposed Negatively Correlated Augmentation (NCA).}
\label{fig:detectionrate}
\end{figure}
We carried out two groups of experiments. In the first group, the model was trained on 68 facial landmarks, and in the second group, we only used very sparse landmarks, to simulate the case of the sheep facial landmarks localisation. The mark-up of sparse landmarks is shown \ref{fig::tri_shape}, which distribute almost uni-formally among the original 68 landmarks on the face. Note that we use the face bounding box detected by dlib face detector \cite{dlib09}, followed by manual check for each face image. This is more realistic in practice than using the tight bounding boxes calculated from the annotated facial landmarks. In order to make a fair comparison, we trained our model as well as most competitive models (highlighted in Section \ref{sec::relatedwork}) including the RCPR, SDM, TREES, ESR, CCNF, LBF, with the same setting. More specifically, we use the same training set and the same bounding box definition. For TCDCN, GNDPM, DRMF we use their initial trained models as their performance is less competitive.

As shown in Fig.~\ref{fig::compare68p}, 1) Our method (NCA + TIF) gets the best performance despite the improvement over the baseline RCRP method is not huge; 2) Only using TIF does not show superior performance over RCPR on dense landmarks setting, which is as expected. The benefit of using TIF is more clear on sparse landmarks, as shown in Fig.~\ref{fig::compare10p}. Our proposed TIF improves the baseline RCPR method as well as the similar ESR method by a large margin. Note that, there are some tricks that are able to make the cascaded pose regression methods more robust such as the smart-restart in \cite{burgos2013robust} and the mirrorability based restart in \cite{yang2015mirror}, which are naturally compatible to our TIF method as well. In this evaluation we are more interested in the benefits brought by the TIF. 


We evaluate the NCA scheme in three methods, our proposed TIF, the RCPR and ESR, since they use the same way of data augmentation.
We set the smallest augmentation number to 11 and the biggest to 40 for the training samples in our NCA method, which makes the total number equal to 20$N$, where $N$ is the number of training samples,  20 is the augmentation number used by the baseline methods. 
In this evaluation, we are more concerned with test samples with big head pose variations. Therefore, we record the successful localisation rates (SLR), i.e. the percentage of test samples are with mean localisation error smaller than 0.1$IOD$. As shown in Fig.\ref{fig:detectionrate}, the proposed NCA scheme is able to improve the SLR effectively. Among the 689 test samples, it is able to successfully localise more than around 30 samples. This is  very significant given the fact that the failures from methods without NCA are already very difficult.

\section{Conclusion and discussion\label{sec::conlusion}}
In this paper, we have addressed the problems of localising key landmarks on sheep and human faces. We proposed a new feature extraction scheme by triplet interpolation(TIF), which is more effective under the conditions of large head pose variation and landmark sparsity. On our new sheep face dataset of only 600 images, our proposed method works considerably well on a large diversity of sheep faces. We also studied the issue of training data imbalance and proposed an sample augmentation strategy to improve the performance on test samples that have big variations.  

Though we have pushed forward the state of the art method for facial landmarks localisation and decreased the failures, there are still failures that are mainly caused by head pose variation or heavy occlusions. It is an open question whether we need to address these challenges explicitly or provide more data similar to the failure cases. Regarding the sheep face analysis, we have only localised the landmarks of interest, there are still many problems to tackle in order to build an automatic computer vision system to identify the sheep in pain. We believe these are all interesting and valuable problems for both computer vision and animal welfare community. 

{\footnotesize 
\bibliographystyle{ieee}
\bibliography{yh}
}

\end{document}